\pdfoutput=1

\documentclass[11pt]{article}
\usepackage[]{acl}
\usepackage{times}
\usepackage{latexsym}
\usepackage{multirow}
\usepackage{graphicx}
\usepackage{caption}
\usepackage{subcaption}
\usepackage{amsmath}
\usepackage{arydshln}
\definecolor{speech}{HTML}{3F6E9E}
\definecolor{text}{HTML}{03820B}
\usepackage[T1]{fontenc}

\usepackage[utf8]{inputenc}

\usepackage{microtype}
\usepackage{fdsymbol}

\title{Attentive Fusion: A Transformer-based Approach to Multimodal Hate Speech Detection}

\author{Atanu Mandal$^{1\dagger}$ \and Gargi Roy$^{2\ddagger}\thanks{\hspace*{0.5em} The work was carried out when the author was at Jadavpur University.}$ \and  Amit Barman$^{3\dagger}$ \\ {\bf Indranil Dutta$^{4\dagger}$} \and {\bf Sudip Kumar Naskar$^{5\dagger}$} \\
        $^{\dagger}$Jadavpur University, Kolkata, INDIA \\ 
        $^{\ddagger}$Optum Global Solutions Private Limited, Bengaluru, INDIA\\ \{$^{1}$atanumandal0491, $^{2}$roygargi1997, $^{3}$amitbarman811, $^{5}$sudip.naskar\}@gmail.com,\\$^{4}$indranildutta.lnl@jadavpuruniversity.in
        } 

\begin{document}
\maketitle
\begin{abstract}
With the recent surge and exponential growth of social media usage, scrutinizing social media content for the presence of any hateful content is of utmost importance. Researchers have been diligently working since the past decade on distinguishing between content that promotes hatred and content that does not. Traditionally, the main focus has been on analyzing textual content. However, recent research attempts have also commenced into the identification of audio-based content. Nevertheless, studies have shown that relying solely on audio or text-based content may be ineffective, as recent upsurge indicates that individuals often employ sarcasm in their speech and writing. To overcome these challenges, we present an approach to identify whether a speech promotes hate or not utilizing both audio and textual representations. Our methodology is based on the Transformer framework that incorporates both audio and text sampling, accompanied by our very own layer called ``Attentive Fusion''. The results of our study surpassed previous state-of-the-art techniques, achieving an impressive macro F1 score of 0.927 on the Test Set.
\end{abstract}

\section{Introduction}
\label{sec: intro}
In recent years, the explosive growth of digital communication platforms has facilitated unprecedented levels of information exchange, enabling individuals from diverse backgrounds to interact and share ideas. However, this surge in online interactions has also led to the emergence of a concerning issue: the increase of hate speech \cite{Davidson_2017}. Hate speech, characterized by offensive, discriminatory, or derogatory language targeting individuals or groups based on race, ethnicity, religion, gender, or sexual orientation, poses significant challenges to maintaining a safe and inclusive online environment \cite{Schmidt_2017}. 

Traditional methods of hate speech detection primarily focused on analyzing text-based content, leveraging natural language processing (NLP) techniques to identify offensive language patterns. While these approaches have yielded some success, they often struggle to capture the nuanced nature of speech, as the exact text might be interpreted differently when considering context, tone, and intent \cite{Fortuna_2018}. To address these limitations, researchers are turning to a more holistic approach combining both text and speech modalities to enhance the accuracy and robustness of hate speech detection systems \cite{Rana_2022}.
\begin{figure}[h!]
\centering
\includegraphics[width=0.3\textwidth,keepaspectratio]{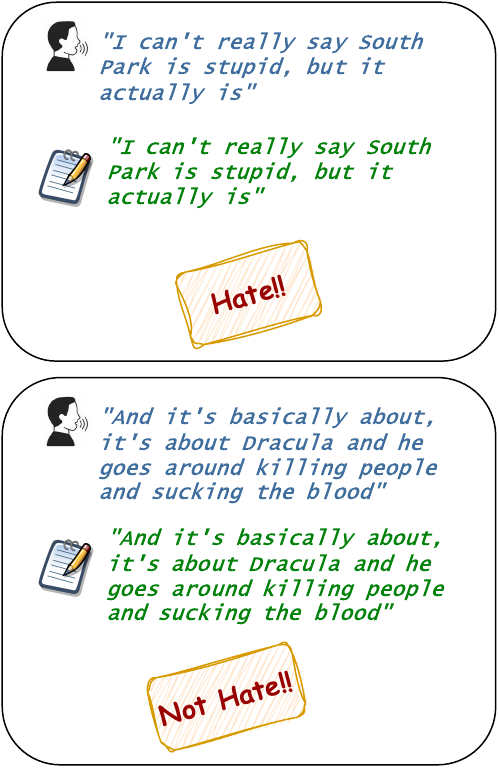}
\caption{Identification of ``Hate'' or ``Not Hate'' using multimodality approach}
\label{fig: multi_approach}
\end{figure}

This multidimensional approach referred to as multimodal hate speech detection, leverages not only the textual content of messages but also the acoustic cues and prosodic features present in speech. By simultaneously analyzing both text and speech-based characteristics, this approach aims to capture a more comprehensive representation of communication, considering not only the words used, but also the emotional nuances conveyed through speech intonation, pitch, and rhythm. Figure \ref{fig: multi_approach} illustrates two examples each for ``Hate'' and ``Not Hate'' using the Multimodality. In the two cues shown (figure \ref{fig: multi_approach}), \colorbox{speech}{ } represents the speech cue and \colorbox{text}{ } represents the text cue.

In this paper, we investigate multimodal hate speech detection exploring the synergies between text and speech for identifying hate speech instances. We examine the challenges posed by hate speech in the digital age, the limitations of traditional text-based detection methods, and the potential advantages of integrating speech data into the detection process. By leveraging insights from various disciplines such as NLP, audio signal processing, and machine learning, multimodal approaches hold promise in achieving higher detection accuracy and reducing false positives, ultimately fostering safer and more inclusive online environments.

Our methodology sets itself apart from other state-of-the-art (SOTA) methodologies in the subsequent manner:
\begin{itemize}
    \vspace{-0.8em}
    \item Our system consists of a sequence of interconnected systems enclosing the Transformer framework\footnote{Code is publicly available in \href{https://github.com/atanumandal0491/Hate-Speech-Identification}{GitHub}.}. 
    \vspace{-0.8em}
    \item We have introduced a layer termed ``Attentive Fusion'' that augments the results.
\end{itemize}

\section{Dataset Description}
\label{sec: dataset_desc}

\begin{table}[h!]
\small
\centering
\begin{tabular}{cccc}
\multirow{2}{*}{\textbf{Dataset}} & \multicolumn{3}{c}{\textbf{Number of Samples}}                    \\ \cline{2-4} 
                                       & \textbf{Train} & \textbf{Dev} & \textbf{Test } \\ \hline \hline
CMU-MOSEI\\[-0.2em] \cite{Mosei2018}    & 597   & 133   & 130   \\ \hline
CMU-MOSI\\[-0.2em] \cite{Mosi2016}     & 181   & 40    & 39    \\ \hline
Common Voice\\[-0.2em] \cite{Common_voice2020} & 8,050 & 1,768 & 1,733 \\ \hline
LJ Speech \\[-0.2em] \cite{ljspeech17}   & 102   & 23    & 23    \\ \hline
MELD\\[-0.2em] \cite{Meld2019}         & 393   & 87    & 85    \\ \hline
Social-IQ\\[-0.2em] \cite{SocialIQ2019}    & 325   & 74    & 69    \\ \hline
VCTK\\[-0.2em] \cite{VCTK2019}         & 138   & 31    & 30    \\ \hline \hline
             & 9,786 & 2,156 & 2,109 \\ 
\end{tabular}
\caption{Statistics of the dataset used for Identification of Hatred.}
\label{tab: Hatred_stat}
\end{table}
For our experiments, we used fragments of the DeToxy dataset \cite{ghosh22}\footnote{\citet{ghosh22} used 20,271 data consisting of CMU-MOSEI, CMU-MOSI, Common Voice, IEMOCAP, LJ Speech, MELD, MSP-Improv, MSP-Podcast, Social-IQ, Switchboard, and VCTK of which IEMOCAP, MSP-Improv, MSP-Podcast, Switchboard are not open-sourced therefore we were unable to use the dataset.}, a dataset for detecting Hatred within spoken English speech. This dataset is derived from diverse open-source datasets. The specifics regarding the number of samples utilized from various datasets are precisely outlined in Table \ref{tab: Hatred_stat}.

\begin{figure}[h!]
\small
     \centering
     \begin{subfigure}[b]{0.45\textwidth}
         \centering
         \includegraphics[width=\textwidth]{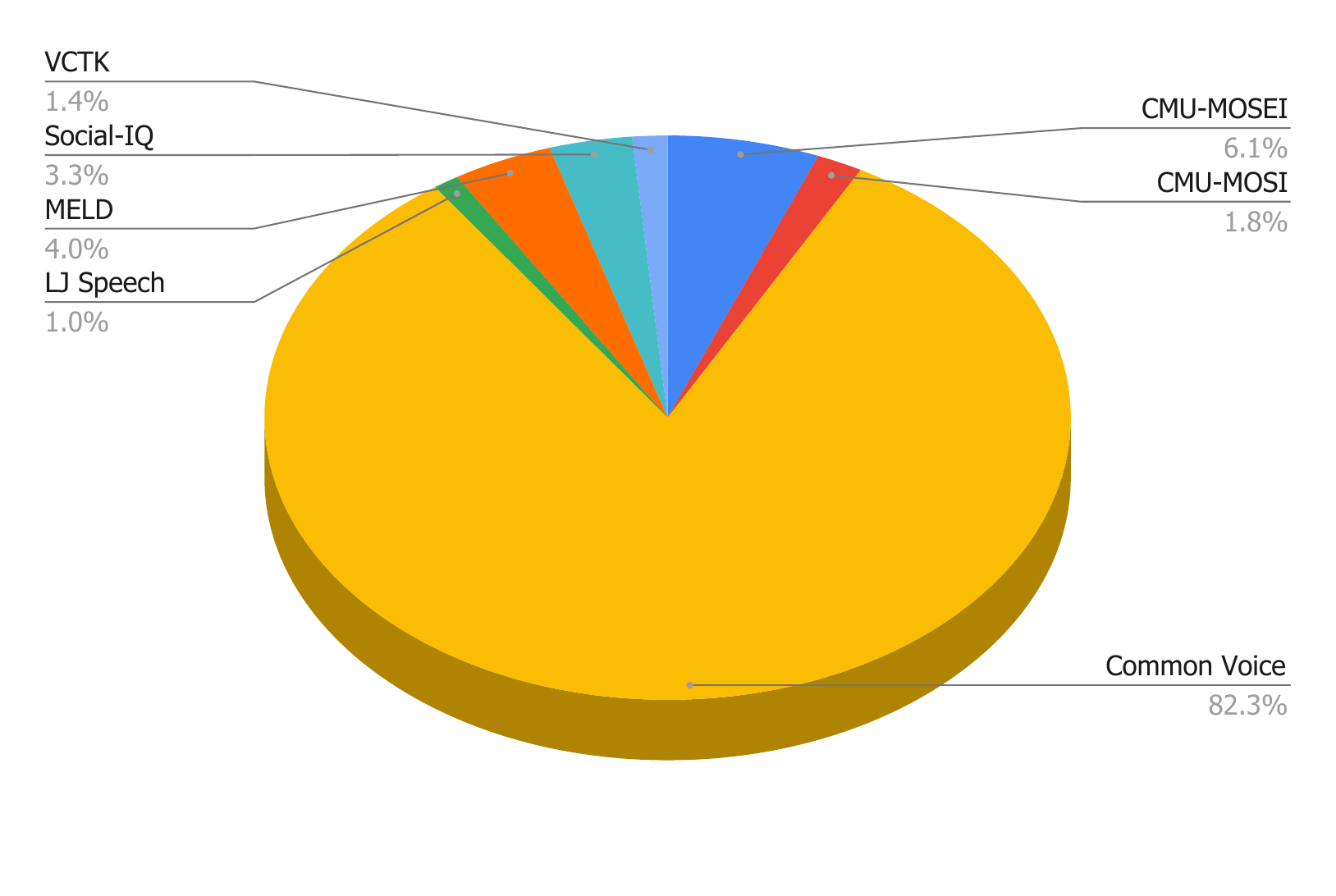}
         \caption{Training Data}
         \label{fig: train_pic_repre}
     \end{subfigure}
     \hfill
     \begin{subfigure}[b]{0.45\textwidth}
         \centering
         \includegraphics[width=\textwidth]{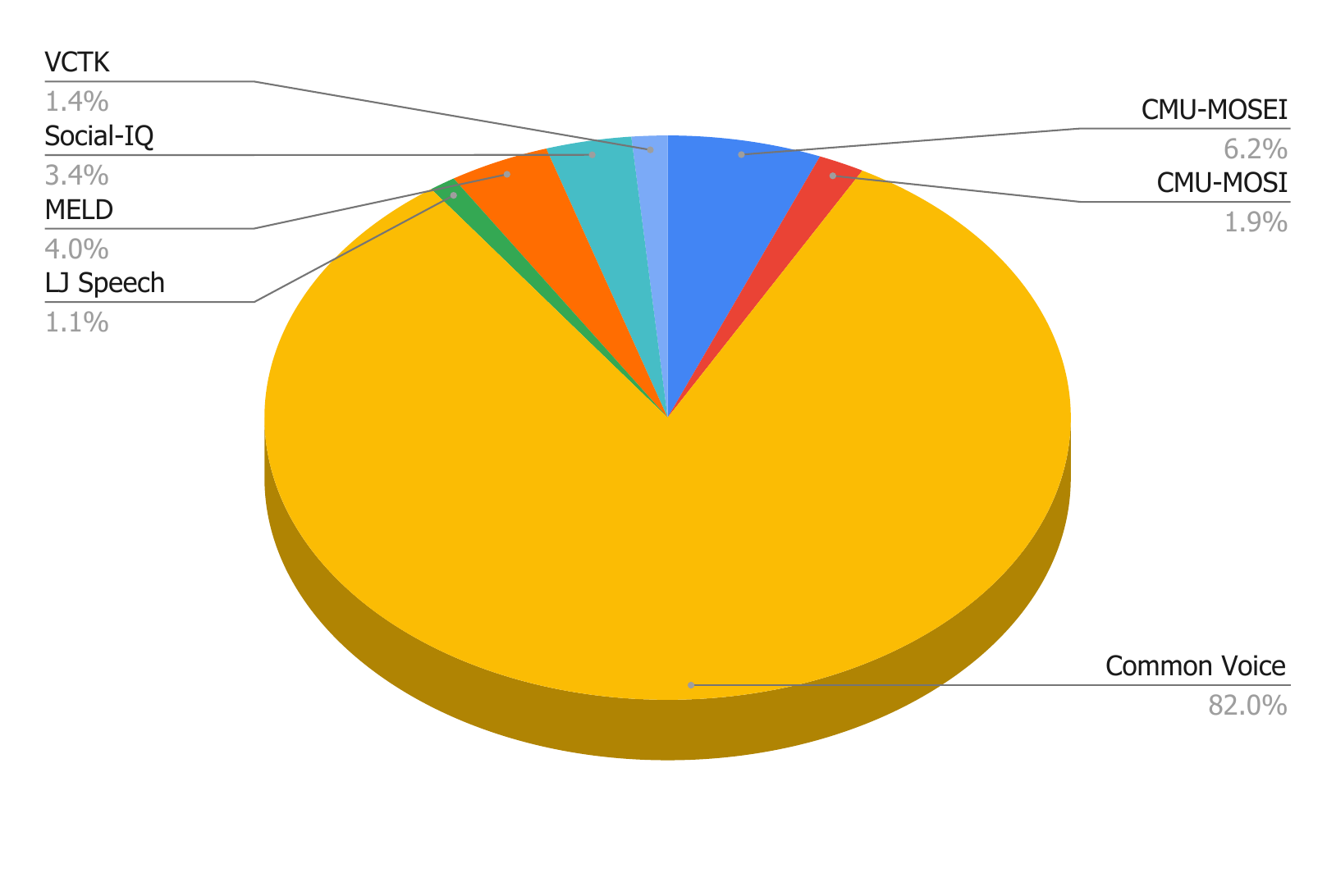}
         \caption{Development Data}
         \label{fig: val_pic_repre}
     \end{subfigure}
     \hfill
     \begin{subfigure}[b]{0.45\textwidth}
         \centering
         \includegraphics[width=\textwidth]{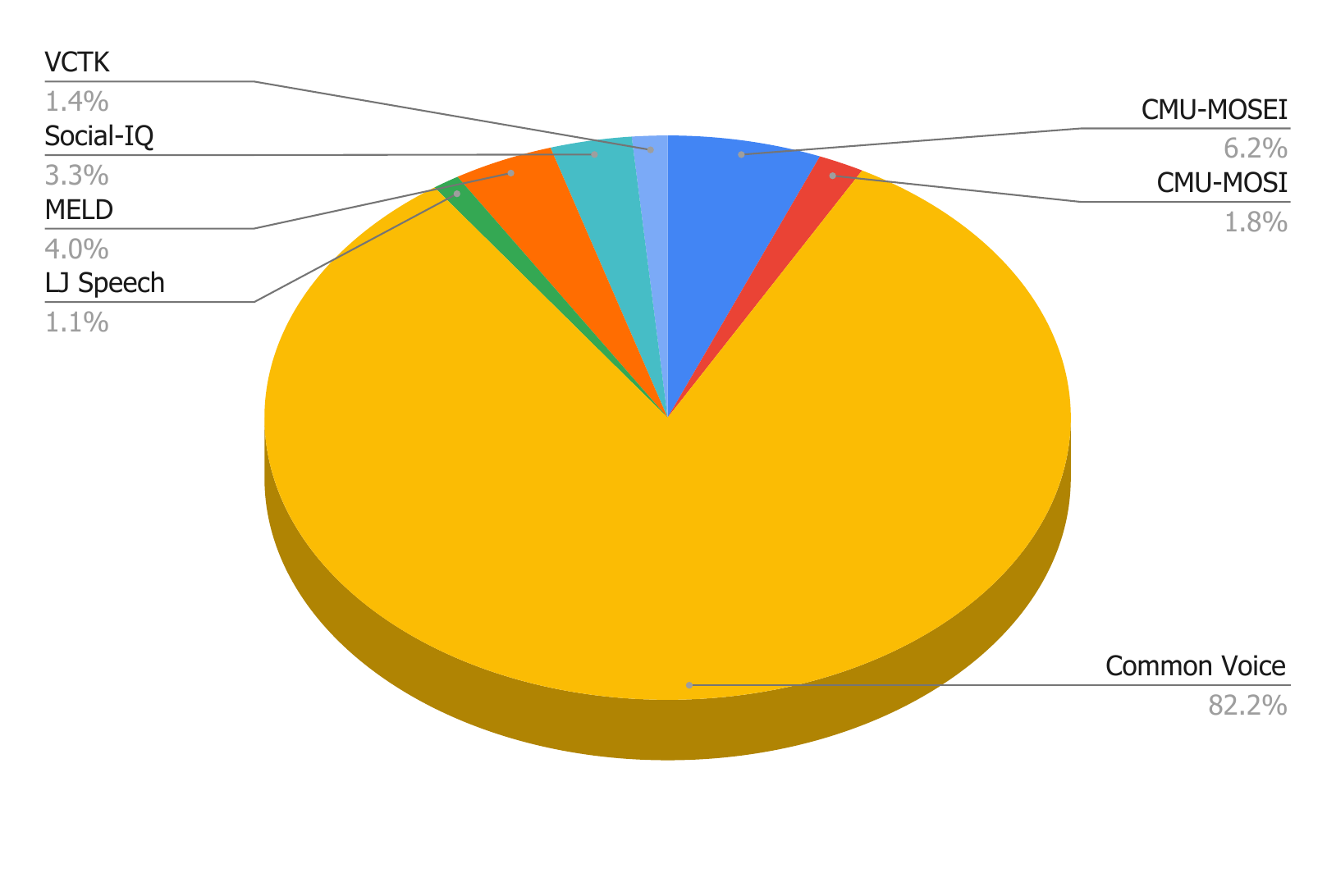}
         \caption{Test Data}
         \label{fig: test_pic_repre}
     \end{subfigure}
        \caption{Pictorial representation of the contribution of datasets}
        \label{fig: pic_repre}
\end{figure}

\begin{table*}[h!]
\centering
\begin{tabular}{c|ccc|ccc}
\textbf{}                                            & \multicolumn{3}{c|}{\textbf{Hate}} & \multicolumn{3}{c}{\textbf{Not Hate}} \\ \cline{2-7}
\textbf{Dataset} &
  \textbf{Train} &
  \textbf{Dev} &
  \textbf{Test} &
  \textbf{Train} &
  \textbf{Dev} &
  \textbf{Test} \\ \hline \hline
CMU-MOSEI & 149         & 33         & 35        & 448        & 100        & 95       \\ \hline
CMU-MOSI &  47        & 10          & 10        &  134        & 30        & 29       \\ \hline
Common Voice &
  2,013 &
  442 &
  433 &
  6,037 &
  1,326 &
  1,300 \\ \hline
LJ Speech & 28          & 6          & 6        &  74        & 17         & 17        \\ \hline
MELD                                                 & 99         & 22          & 21        &  294        & 65        & 64       \\ \hline
Social-IQ                                            &  83        &  18         & 19        &  242        &   56      & 50       \\ \hline
VCTK                                                 &  34        & 8          & 8        &  104        & 23         & 22  \\ \hline \hline
 & 2,453 & 539 & 532 & 7,333 & 1,617 & 1,577  \\
\end{tabular}
\caption{Data Statistics of ``Hate'' and ``Not Hate'' }
\label{tab: overall_stat}
\end{table*}

\begin{figure}[h!]
    \centering\includegraphics[width=0.4\textwidth,keepaspectratio]{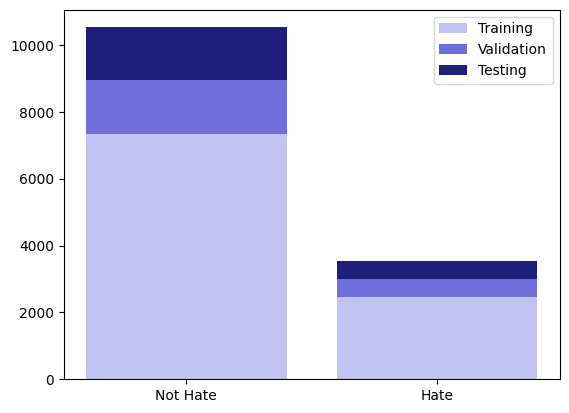}
    \caption{Sample count for ``Hate'' and ``Not Hate''}
    \label{fig: overall_bar}
\end{figure}
Our experiments were carried out on a comprehensive dataset that encompassed all seven datasets combined. Each dataset contained entries that fell into either the "Hate" or "Not-Hate" category, along with a transcription for each audio. To facilitate understanding, we have depicted the distribution of each dataset's contribution to our framework through a pie chart, as showcased in Figure \ref{fig: pic_repre}. Figures \ref{fig: train_pic_repre}, \ref{fig: val_pic_repre}, and \ref{fig: test_pic_repre} accordingly illustrate the respective contributions of the training data, development data, and testing data. There exists a significant disparity in the number of samples across the various datasets but, the proportional representation of the training, development, and test datasets remains consistent. Notably, Common Voice comprises the majority of the data, while LJ Speech is the least represented. The statistical analysis of the ``Hate'' and ``Not Hate'' classes is presented in Table \ref{tab: overall_stat}. Meanwhile, the bar plot showcasing the sample count for both classes can be seen in Figure \ref{fig: overall_bar}. A comprehensive description of datasets is described in Section \ref{subsec: mosei}, \ref{subsec: mosi}, \ref{subsec: cv}, \ref{subsec: lj}, \ref{subsec: meld}, \ref{subsec: social}, and \ref{subsec: vctk}.

\begin{figure*}[h!]
     \centering
      \begin{subfigure}[b]{0.225\textwidth}
         \centering
         \includegraphics[width=\textwidth]{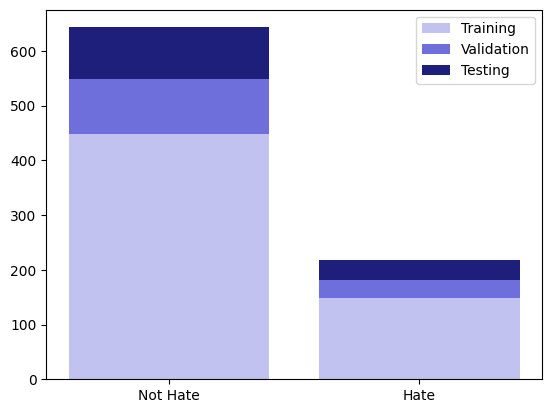}
         \caption{CMU-MOSEI}
         \label{fig: mosei_bar}
     \end{subfigure}
     \hfill
     \begin{subfigure}[b]{0.225\textwidth}
         \centering
         \includegraphics[width=\textwidth]{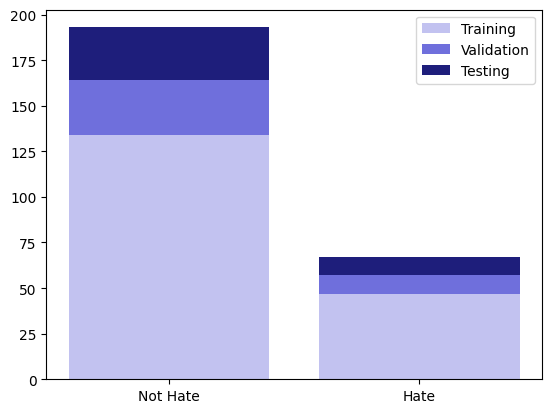}
         \caption{CMU-MOSI}
         \label{fig: mosi_bar}
     \end{subfigure}
     \hfill
      \begin{subfigure}[b]{0.225\textwidth}
         \centering
         \includegraphics[width=\textwidth]{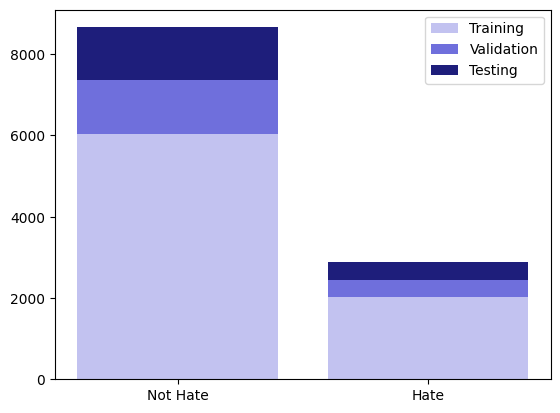}
         \caption{Common Voice}
         \label{fig: cv_bar}
     \end{subfigure}
     \hfill
     \begin{subfigure}[b]{0.225\textwidth}
         \centering
         \includegraphics[width=\textwidth]{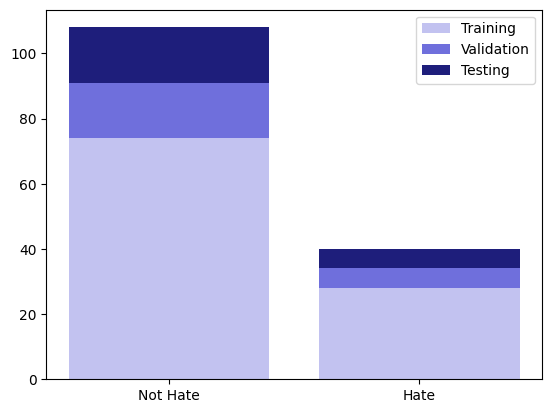}
         \caption{LJ Speech}
         \label{fig: lj_bar}
     \end{subfigure}
     \hfill
     \begin{subfigure}[b]{0.225\textwidth}
         \centering
         \includegraphics[width=\textwidth]{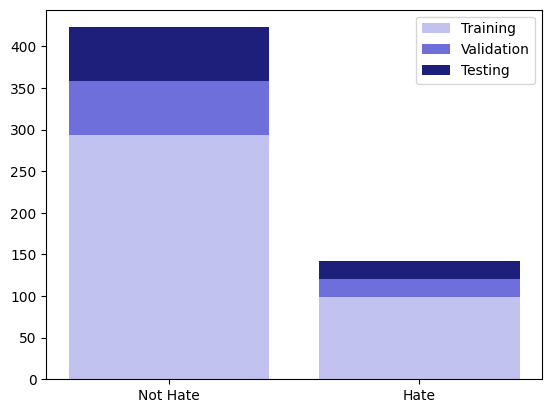}
         \caption{MELD}
         \label{fig: meld_bar}
     \end{subfigure}
     \hfill
     \begin{subfigure}[b]{0.225\textwidth}
         \centering
         \includegraphics[width=\textwidth]{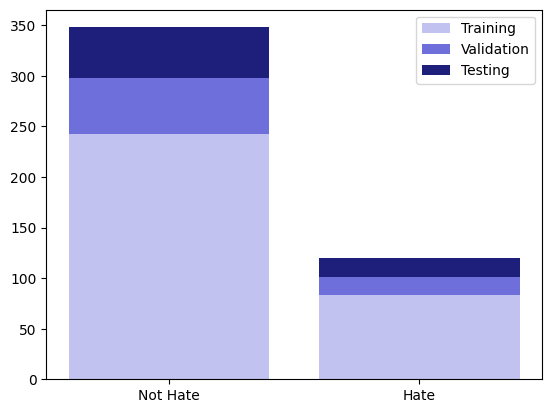}
         \caption{Social-IQ}
         \label{fig: social_bar}
     \end{subfigure}
     \hfill
     \begin{subfigure}[b]{0.225\textwidth}
         \centering
         \includegraphics[width=\textwidth]{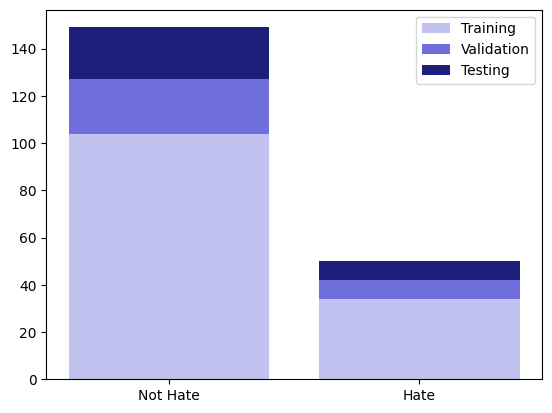}
         \caption{VCTK}
         \label{fig: vctk_bar}
     \end{subfigure}
        \caption{Sample count for ``Hate'' and ``Not Hate''}
        \label{fig: bar_repre}
\end{figure*}

\begin{figure*}[h!]
     \centering
      \begin{subfigure}[b]{0.225\textwidth}
         \centering
         \includegraphics[width=\textwidth]{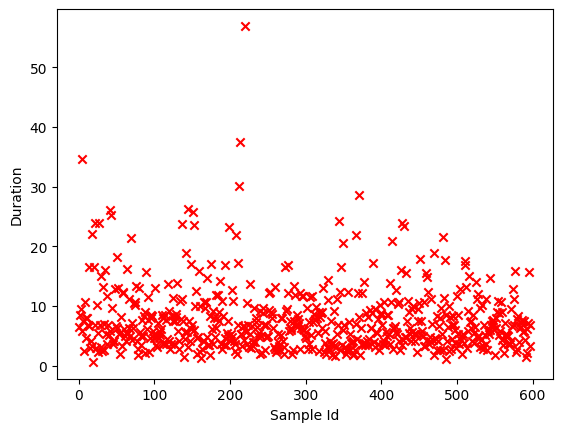}
         \caption{CMU-MOSEI}
         \label{fig: mosei_scatter}
     \end{subfigure}
     \hfill
     \begin{subfigure}[b]{0.225\textwidth}
         \centering
         \includegraphics[width=\textwidth]{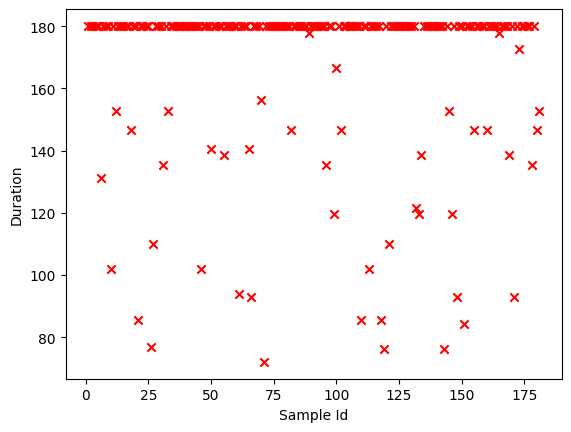}
         \caption{CMU-MOSI}
         \label{fig: mosi_scatter}
     \end{subfigure}
     \hfill
      \begin{subfigure}[b]{0.225\textwidth}
         \centering
         \includegraphics[width=\textwidth]{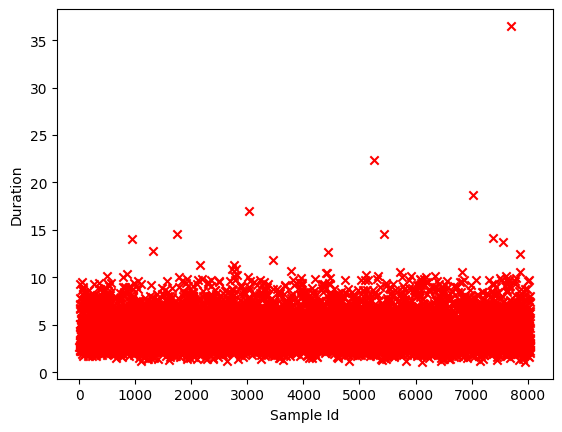}
         \caption{Common Voice}
         \label{fig: cv_scatter}
     \end{subfigure}
     \hfill
     \begin{subfigure}[b]{0.225\textwidth}
         \centering
         \includegraphics[width=\textwidth]{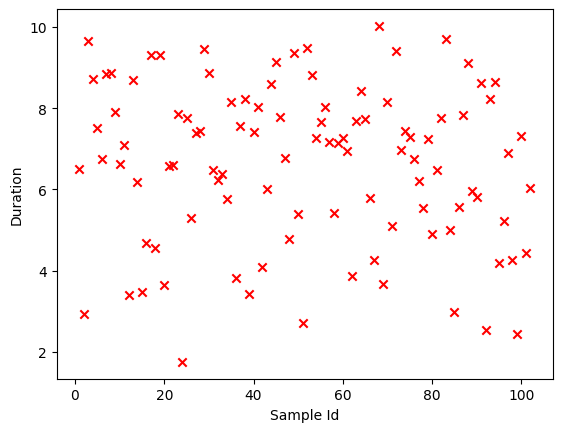}
         \caption{LJ Speech}
         \label{fig: lj_scatter}
     \end{subfigure}
     \hfill
     \begin{subfigure}[b]{0.225\textwidth}
         \centering
         \includegraphics[width=\textwidth]{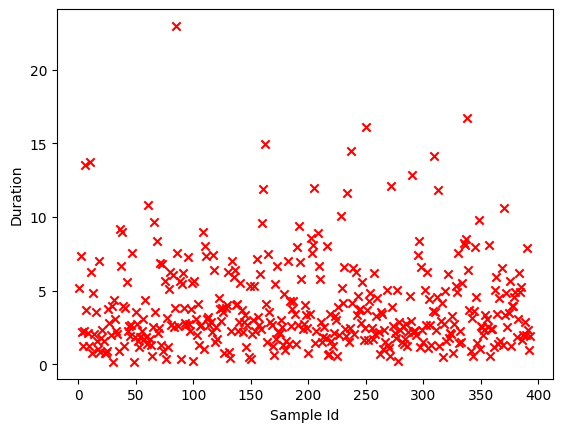}
         \caption{MELD}
         \label{fig: meld_scatter}
     \end{subfigure}
     \hfill
     \begin{subfigure}[b]{0.225\textwidth}
         \centering
         \includegraphics[width=\textwidth]{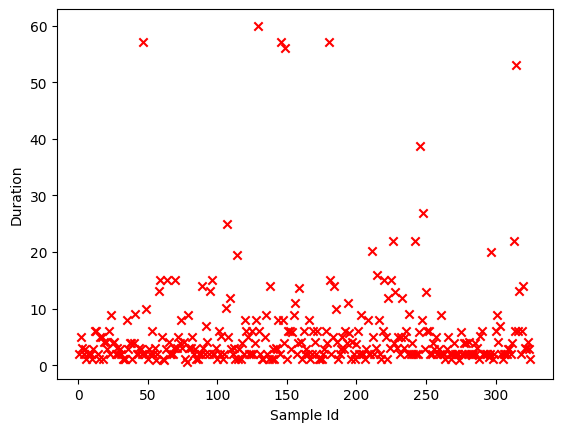}
         \caption{Social-IQ}
         \label{fig: social_scatter}
     \end{subfigure}
     \hfill
     \begin{subfigure}[b]{0.225\textwidth}
         \centering
         \includegraphics[width=\textwidth]{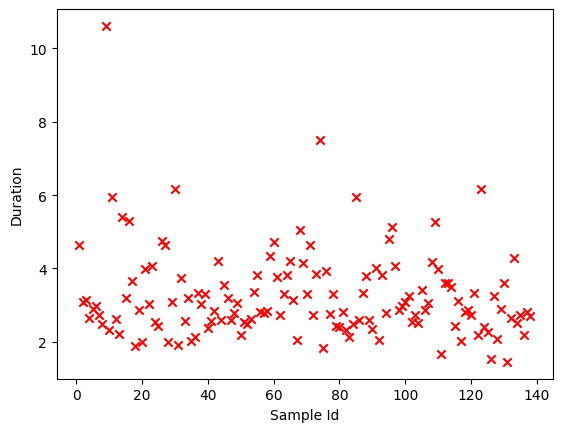}
         \caption{VCTK}
         \label{fig: vctk_scatter}
     \end{subfigure}
        \caption{Scatter representation of Datasets according to audio length}
        \label{fig: scatter_repre}
\end{figure*}

\subsection{CMU-MOSEI}
\label{subsec: mosei}
Carnegie Mellon University - Multimodal Opinion Sentiment and Emotion Intensity (CMU-MOSEI) \cite{Mosei2018} is considered the largest and most extensive dataset for emotion recognition tasks and multimodal sentiment analysis. Figure \ref{fig: mosei_bar} provides the information on the number of samples for ``Hate'' and ``Not Hate'' and Figure \ref{fig: mosei_scatter} provides the pictorial information of the number of samples to audio duration.

\subsection{CMU-MOSI}
\label{subsec: mosi}
Carnegie Mellon University - Multimodal Corpus of Sentiment Intensity (CMU-MOSI) \cite{Mosi2016} is another dataset by Carnegie Mellon University, which consists of 2199 video clips of different opinions, annotated with sentiment. It is annotated in the range $[-3,3]$, using various parameters for sentiment intensity, subjectivity, and per-millisecond annotations of audio features. It contains 97\% non-toxic and nearly 3\% toxic utterances. Figure \ref{fig: mosi_bar} provides the information on the number of samples for ``Hate'' and ``Not Hate'' and Figure \ref{fig: mosi_scatter} provides the pictorial information of the number of samples to audio duration.

\subsection{Common Voice}
\label{subsec: cv}
This dataset \cite{Common_voice2020} by Mozilla Developer Network is an open-source, dataset of voices of multiple languages for the use of training speech-enabled systems, with 20,217 hours of recorded audio and 14,973 hours of validated speech audios. Figure \ref{fig: cv_bar} provides the information on the number of samples for ``Hate'' and ``Not Hate'' and Figure \ref{fig: cv_scatter} provides the pictorial information of the number of samples to audio duration.

\subsection{LJ Speech}
\label{subsec: lj}
This is another open-source dataset \cite{ljspeech17} which has 13,100 clips of short audio segments, with one speaker reading texts from a collection of seven books of non-fiction. Every clip is transcribed and has a varying length of 1 to 10 seconds. Figure \ref{fig: lj_bar} provides the information on the number of samples for ``Hate'' and ``Not Hate'' and Figure \ref{fig: lj_scatter} provides the pictorial information of the number of samples to audio duration. 

\vspace{-0.3em}

\subsection{MELD}
\label{subsec: meld}
Multimodal Emotion Lines Dataset (MELD) \cite{Meld2019} has over 1,400 dialogues and 13,000 dialogues from the television show ``Friends''. Each utterance in dialogue has been labelled by one of the emotions -- Anger, Disgust, Sadness, Joy, Neutral, Surprise, and Fear. MELD also has annotations for sentiments -- positive, negative, and neutral. Figure \ref{fig: meld_bar} provides the information on the number of samples for ``Hate'' and ``Not Hate'' and Figure \ref{fig: meld_scatter} provides the pictorial information of the number of samples to audio duration.

\subsection{Social-IQ}
\label{subsec: social}
Another dataset \cite{SocialIQ2019} by Carnegie Mellon University has videos that are thoroughly validated and annotated, along with questions, answers, and annotations for the level of complexity of the said questions and answers. Figure \ref{fig: social_bar} provides the information on the number of samples for ``Hate'' and ``Not Hate'' and Figure \ref{fig: social_scatter} provides the pictorial information of the number of samples to audio duration.

\subsection{VCTK}
\label{subsec: vctk}
The VCTK corpus \cite{VCTK2019} contains 110 speakers' speech data spoken in English, having various accents. Every single speaker reads a passage, selected from newspapers, archives, and so on. Figure \ref{fig: vctk_bar} provides the information on the number of samples for ``Hate'' and ``Not Hate'' and Figure \ref{fig: vctk_scatter} provides the pictorial information of the number of samples to audio duration. 

\section{Experiments}
\label{sec: e&r}
This section demonstrates our innovative techniques for detecting Hatred within a speech. The section is divided into numerous subsections for understanding our approach with ease. Section \ref{subsec: dataset_pre_processing} presents the methods we used to prepare the dataset for our suggested framework. Section \ref{subsec: framework} describes our suggested framework. Section \ref{subsec: hyper} discusses the parameters used for our proposed framework and Section \ref{subsec: results} discusses the results of our approach with other benchmark frameworks.

\subsection{Dataset Pre-processing}
\label{subsec: dataset_pre_processing}
In the task of pre-processing, we carefully selected the data that possessed comparable lengths of audio. We disregarded instances with excessively long or short duration. Our inclination to overlook excessively long audio duration stemmed from the understanding that it would necessitate extensive computational resources. Conversely, audio with extremely short duration lacked the richness of audio features.

\begin{table}[h!]
\centering
\begin{tabular}{lc}
                            & Values   \\ \hline \hline
Sample Rate                 & 16,000 Hz   \\ \hline
Number of FFT               & 400      \\ \hline
Number of MELs Channel      & 80       \\ \hline
Hop Length                  & 160      \\ \hline
Chunk Length                & 30       \\ \hline
Number of Samples           & 4,80,000 \\ \hline
Number of Frames            & 3,000    \\ \hline
Number of Samples per Token & 320      \\ \hline
Frames per Second           & 10 ms    \\ \hline
Tokens per Second           & 25 ms   \\ \hline
\end{tabular}
\caption{Audio feature extraction parameters}
\label{tab: feature_hyper}
\end{table}

We conducted a series of experiments with our framework, exploring various methods of extracting features such as Mel-frequency cepstral coefficients (MFCCs) and filter banks. However, we discovered that the ``log mel spectrogram'' yields superior accuracy in comparison to other alternatives, as it captures auditory information in a manner akin to human perception. To extract these features, we established the optimal parameters empirically, which are detailed in Table \ref{tab: feature_hyper}.

For feature extraction from text, we used the pre-trained Albert Tokenizer \cite{AlBert_2020} from IndicBART \cite{indicbart_2022} developed by AI4Bharat. To tokenize each sentence, we introduced the symbols ``$<s>$'' and ``$</s>$'' at the start and end, respectively, signifying the commencement and conclusion of the sentences.

\subsection{Framework}
\label{subsec: framework}
\begin{figure*}[h!]
\centering
\includegraphics[width=\textwidth,keepaspectratio]{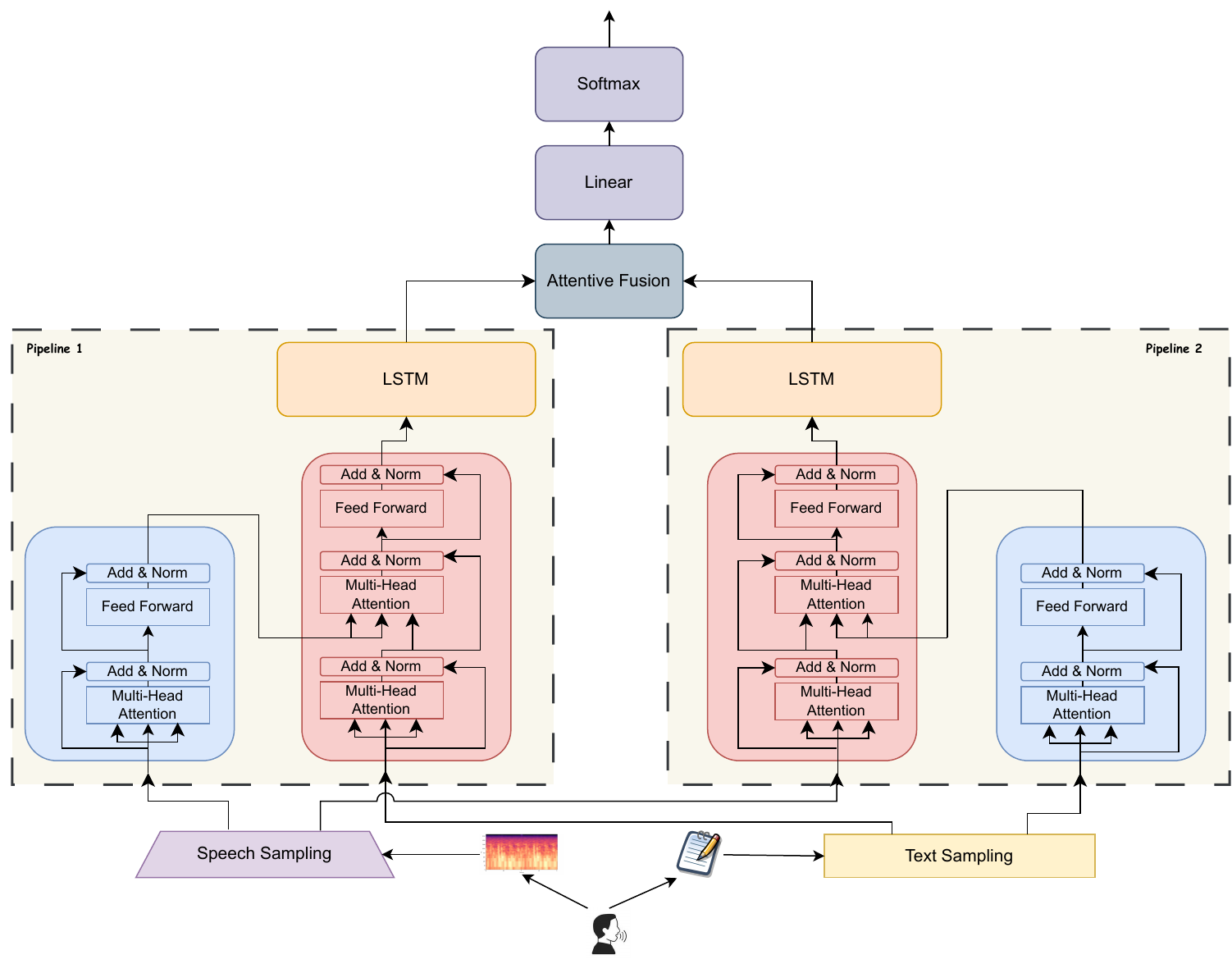}
\caption{Overview of our approach}
\label{fig: framework}
\end{figure*}

We have used the Transformer \cite{Vaswani_2017} 
framework which has gained widespread recognition and is considered SOTA in the domains of Speech Recognition and Machine Translation (MT) 
due to its exceptional ability to handle the complexities of these complex tasks. To provide a clear overview of our methodology, Figure \ref{fig: framework} presents an overview of our framework.

The Speech Feature is extracted by the ``log mel spectrogram'' technique, which has been discussed in section \ref{subsec: dataset_pre_processing}. This technique involves the computation of a spectrogram that represents the frequency content of an audio signal over time, using a logarithmic scale for the frequency axis. The resulting spectrogram has a dimension of ``$(80 \times time\_step)$'' and is then passed to the Speech Sampling Block. The Speech Sampling Block is responsible for selecting a subset of the input spectrogram, based on certain criteria (described in Section \ref{subsubsec: audio_sample}). On the other hand, the tokenized Text, which is obtained through a process described in section \ref{subsec: dataset_pre_processing}, has a dimension of ``$(max\_length \times 1)$'' and is passed to the Text Sampling Block (discussed in Section \ref{subsubsec: text_sample}). The Text Sampling Block performs a similar function as the Speech Sampling Block but on the tokenized Text instead of the spectrogram. 

The resulting subset of Speech Sampling is fed to the Encoder of the first Transformer module and the Decoder of the second Transformer module. Similarly, the resulting subset of Text Sampling is fed to the Decoder of the first Transformer module and the Encoder of the second Transformer module. The motivation behind this approach is to investigate whether the text in the Decoder can learn from the audio in the Encoder, and vice versa. This is inspired by the idea of MT, where the target text in the Decoder learns from the source text in the Encoder. By applying this concept to the audio and text domains, we aim to explore the potential for cross-modal learning and the transfer of knowledge between different modalities.

To further process the outputs of the two Transformer modules, we introduce a Long short-term memory (LSTM) block that consists of a single LSTM layer. This LSTM block is responsible for sequentially learning the knowledge from each step of the output. After going through this process, we obtain two outputs: one from the first LSTM and another from the second LSTM. The combination of the first Transformer with the first LSTM is represented as ``Pipeline 1'' and the combination of the second Transformer with the second LSTM is represented as ``Pipeline 2''. These two outputs are then passed to the proposed ``Attentive Fusion'' Layer (described in Section \ref{subsubsec: atten_layer}). The Attentive Fusion Layer is designed to learn the knowledge from both outputs in a joint manner, combining the information from the two pipelines. The output of the Attentive Fusion Layer is then fed to a Linear Layer with Softmax activation, where it undergoes further processing and classification according to the specific classes. 

A comprehensive exploration of the Audio Sampling Module, Text Sampling Module, and our proposed Attentive Fusion layer are presented in Section \ref{subsubsec: audio_sample}, \ref{subsubsec: text_sample}, and \ref{subsubsec: atten_layer}, respectively. All hyperparameter configurations can be found in Section \ref{subsec: hyper}.

\subsubsection{Speech Sampling}
\label{subsubsec: audio_sample}
\begin{figure}[h!]
\centering
\includegraphics[width=0.25\textwidth,keepaspectratio]{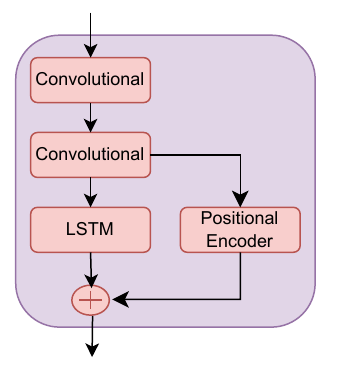}
\caption{Overview of Speech Sampling}
\label{fig: speech_sampling}
\end{figure}
Our module for ``Speech Sampling'' was influenced by the work of \citet{whisper_2022} with minor modifications. This module comprises a pair of Convolutional layers, with each layer being accompanied by a Gaussian Error Linear Unit (GELU) activation function. The outcome of the Convolutional layer was passed through a Positional Encoder and an LSTM layer separately. The results from the Positional Encoder and LSTM were combined. The Speech Sampling framework is shown in Figure \ref{fig: speech_sampling}.

\subsubsection{Text Sampling}
\label{subsubsec: text_sample}

\begin{figure}[h!]
\centering
\includegraphics[width=0.2\textwidth,keepaspectratio]{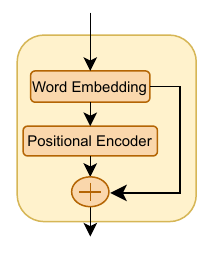}
\caption{Overview of Text Sampling}
\label{fig: text_sampling}
\end{figure}

Our ``Text Sampling'' module comprises a simplistic approach containing Word Embedding and a Positional Encoder. The raw text was tokenized by appending with ``$<s>$'' and ``$</s>$'' at the commencement and conclusion of the sentences (refer to Section \ref{subsec: dataset_pre_processing}) and passed on to Word Embedding. The subsequent output is then directed to the Positional Encoder. Subsequently, the output of the Word Embedding and the Positional Encoder are combined and conveyed to the subsequent hierarchical module. The representation of the ``Text Sampling'' framework can be seen in Figure \ref{fig: text_sampling}.

\subsubsection{Attentive Fusion Layer}
\label{subsubsec: atten_layer}

The layer we have named the ``Attentive Fusion'' layer is a layer that we have devised for the purpose of detecting hatred within a speech. In our methodology (as illustrated in equation \ref{eq: atten_layer_1}), we have seamlessly integrated the outcomes from Pipeline 1 and Pipeline 2, allowing them to flow into their respective Linear layers individually, thereby ensuring the preservation of their unique characteristics.

\begin{equation}
\label{eq: atten_layer_1}
\begin{aligned}
    L_{1} = Linear(x_{1})\\
    L_{2} = Linear(x_{2})\\
\end{aligned}
\end{equation}

The result, $L_{1}$ and $L_{2}$ underwent a process of cross multiplication, after which a hyperbolic tangent function ($tanh$) was used. To enhance the disparity of each tensor value ($w_{i}$), an exponential ($e$) function was applied, as demonstrated in equation \ref{eq: atten_layer_2}.

\begin{equation}
\label{eq: atten_layer_2}
    w_{i} = e^{(tanh(L_{1} \times L_{2}))}
\end{equation}

The outcome of equation \ref{eq: atten_layer_2} $w_{i}$ underwent division by the summation of every element of $w_{i}$. To prevent division by zero, we introduced an epsilon ($\epsilon$). The entire outcome was then subjected to multiplication with $w_{i}$ itself. Equation \ref{eq: atten_layer_3} illustrates our approach.

\begin{equation}
\label{eq: atten_layer_3}
    w^{'}_{i} = \frac{w_{i}}{\sum_{i}w_{i} + \epsilon} \times w_{i}
\end{equation}

The value, $ w^{'}_{i} $ obtained from equation \ref{eq: atten_layer_3} was introduced into the subsequent module that incorporates a Linear Layer to differentiate between different classes.

\subsection{Hyperparameters}
\label{subsec: hyper}

\subsubsection{Speech Sampling}
For the two Convolutional layers, we used filter sizes of ``4096'' and ``1024'', respectively and kernel size of ``3'' for both. Strides of ``1'' for $1^{st}$ and ``2'' for $2^{nd}$ Convolutional layer was used. For LSTM layer units of ``512'' with activation function ``tanh'' and recurrent activation of ``sigmoid'' was used. For the Positional Encoder vocab size of ``64,014'', the hidden dimension of ``512'' was passed. 

\subsubsection{Text Sampling}
For the Word Embedding, we used a vocab size of ``64,014'', and a hidden dimension of ``512'' with True mask zero. For the Positional Encoder vocab size of ``64,014'', the hidden dimension of ``512'' was passed.

\subsubsection{Transformer}
For the transformer framework, the number of heads and the number of layers for the Encoder and Decoder were kept ``4''. The hidden dimensions were kept ``512'', and the dropout \cite{Srivastava_2014} rate of ``0.3''.

\subsubsection{LSTM}
For the LSTM layer, units of ``512'' with activation function ``tanh'' and recurrent activation of ``sigmoid'' were used, with a dropout rate of ``0.3''. Use of bias and Forget bias with return sequence kept True. 

\subsubsection{Learning Rate}
We use ``AdamW'' optimizer \cite{Ilya_2019} with $\beta_{1}$ of ``0.9'', $\beta_{2}$ of ``0.98'', $\epsilon$ of ``$1 \times 10^{-6}$'' and decay of ``0.1'' with adaptive learning rate.

\begin{equation}
    \label{eq: lrate}
    \begin{aligned}
        &arg_{1} = \sqrt{cs} \\
        &arg_{2} = cs \times ws^{-1.5} \\
        &lr = \sqrt{d_{model}} \times \min(arg_{1}, arg_{2}) \\
        &lrate = \min(lr, 0.0004) \\
    \end{aligned}
\end{equation}

In equation \ref{eq: lrate}, $cs$ is current step, warmup steps ($ws$) were set to ``2048'' and $d_{model}$ of ``512''.

\section{Results}
\label{subsec: results}
Table \ref{tab: er_diff_system} provides the benchmark macro F1 scores of the frameworks proposed by \citet{ghosh22}. The proposed frameworks were trained only with the audio sequences. Our study suggests that only the audio sequences cannot provide a better understanding of hatred in speech. Current trends show that persons use hateful words in spoken sentences but the tones, frequency and amplitudes are kept normal, which can also be remarked as ``Sarcasmic Behaviour''. To overcome the situation we used multimodality where an audio specimen along with its transcripts are used. Using multimodality has an increase in macro F1 score compared with the F-Bank framework proposed by \citet{ghosh22}. 

In cases of unfrozen wav2vec-2.0. the differences are very nominal as wav2vec-2.0 provides an embedding knowledge of each token of the audio specimen. In contrast, we didn't use any embedding knowledge of speech tokens, which will be experimented with in our upcoming works. The researcher has shown two wav2vec-2.0 among which is one wav2vec-2.0 (9 layer). In this system, the researcher took the representation token from the $9^{th}$ layer.

\begin{table}[h!]
\centering
\begin{tabular}{cccc}
\textbf{System}                       & \textbf{Category} & \textbf{Dev} & \textbf{Test} \\ \hline \hline
F-Bank                       & -                 & 0.610        & 0.620         \\ \hline
\multirow{2}{*}{wav2vec-2.0} & Freezed           & 0.448        & 0.457         \\ \cline{2-4}
                                      & Unfreezed         & 0.877        & 0.869         \\ \hline
\begin{tabular}[c]{@{}c@{}}wav2vec-2.0\\[-0.3em]  (9 layer)\end{tabular} & Unfreezed                                                      & 0.897 & 0.877 \\ \hline
\begin{tabular}[c]{@{}c@{}}Proposed \\[-0.3em] Framework\end{tabular}    & - &  \textbf{0.931} & \textbf{0.927}   \\ \hline  
\end{tabular}
\caption{Evaluation and Result of Different Systems Proposed by \citet{ghosh22}. \textit{Unfrozen wav2vec-2.0 setup with representations taken from the $9^{th}$ layer as reported by \citet{ghosh22}}}
\label{tab: er_diff_system}
\end{table}

\section{Ablation Studies}
\subsection{``Attentive Fusion layer'' vs ``Concatenate layer''}
Instead of using our proposed Attentive Fusion Layer, we used the Concatenate Layer to study the effectiveness of the Attentive Fusion layer and found it even outperformed the Concatenate layer. The differences in results are shown in Table \ref{tab: abl_1}.

\begin{table}[h!]
\centering
\begin{tabular}{ccc}
                           & \textbf{Dev} & \textbf{Test} \\ \hline \hline
Concatenate Layer & 0.908          & 0.909         \\ \hline
Attentive Fusion Layer   & \textbf{0.931}          & \textbf{0.927}         \\ \hline
\end{tabular}
\caption{Macro F1 Score Result on replacing ``Attentive Fusion layer'' with ``Concatenate layer''}
\label{tab: abl_1}
\end{table}

\subsection{Using Pipeline 1 and Pipeline 2 separately}

We also checked whether alone text learning from audio representation or audio learning from textual representation outperformed our baseline result or not. For the evaluation, we used Pipeline 1 and Pipeline 2 separately followed by Linear Layer, and found that it is unable to score at par result to our baseline. The differences in results are shown in Table \ref{tab: abl_2}.

\begin{table}[h!]
\centering
\begin{tabular}{ccc}
                    & \textbf{Dev} & \textbf{Test} \\ \hline \hline
Pipeline 1 & 0.910          & 0.909         \\ \hline
Pipeline 2 & 0.910          & 0.899         \\ \hline
\multicolumn{1}{l}{Our Baseline} & \multicolumn{1}{l}{\textbf{0.931}} & \multicolumn{1}{l}{\textbf{0.927}} \\ \hline
\end{tabular}
\caption{Macro F1 Score Result on Pipeline 1 and Pipeline 2 separately and compared with our baseline.}
\label{tab: abl_2}
\end{table}

\section{Conclusion}
In this work, we proposed a framework that can classify whether a speech promotes Hatred or not. For the speech feature extraction, we used a log mel spectrogram feature extraction technique. Our framework consists of Speech Sampling and Text sampling followed by two separate transformer frameworks that serve different efforts. Each Transformer framework is followed by an LSTM layer, the output of which is fed to our proposed layer, and further sent to Linear Layer for Classification. The whole framework was able to outperform the existing benchmark macro F1 score. The only limitation of our approach is it is limited to the English language. In future work, we would like to test its robustness for other languages.

\section*{Acknowledgements}
This research was supported by the TPU Research Cloud (TRC) program, a Google Research initiative.

\bibliography{acl_latex}
\bibliographystyle{acl_natbib}




\end{document}